\documentclass[runningheads]{llncs}

\usepackage{eccv}

\usepackage{eccvabbrv}

\usepackage{graphicx}
\usepackage{booktabs}
\usepackage{float}

\usepackage[accsupp]{axessibility}  

\usepackage{multirow}
\usepackage{siunitx}    

\usepackage{hyperref}

\usepackage{orcidlink}

\begin{document}

\title{BioDCASE: Active Learning for Bioacoustics} 

\author{Ben McEwen\inst{1}\orcidlink{0000-0002-0869-7717} \and
Rupa Kurinchi-Vendhan\inst{2}\orcidlink{0000-0001-9397-6050} \and
Shiqi Zhang\inst{3}\orcidlink{0009-0008-6012-8088}
Lukas Rauch\inst{4}\orcidlink{0000-0002-6552-3270}
Marek Herde\inst{5}
Sara Beery\inst{2}\orcidlink{0000-0002-2544-1844
}}

\authorrunning{B.~McEwen et al.}

\institute{University of Amsterdam, Amsterdam, Netherlands \and Massachusetts Institute of Technology, Cambridge, Massachusetts, USA \and Tampere University, Finland \and Earth Species Project \and Kassel University, Germany}

\maketitle

\begin{abstract}
Ecological monitoring increasingly relies on machine learning models, whose performance depends on the quality and quantity of labelled data. However, obtaining these labels is costly, particularly in passive acoustic monitoring, where vast amounts of data are collected but only a small proportion can feasibly be annotated. Active learning addresses this bottleneck by prioritizing which samples should be labelled. However, progress is difficult to measure, because published methods are evaluated under different models, budgets, evaluation metrics and datasets. To address this challenge, we present the 2026 Active Learning for Bioacoustics BioDCASE challenge: a systematic evaluation of sampling methods designed to identify effective AL strategies. Participant methods were evaluated across four subsets composed of terrestrial and marine data. Across ten proposed sampling methods from seven teams, the top-ranked method achieved an area under the learning curve \qty{26.4}{\percent} higher than random sampling at the same annotation budget, averaged over four data subsets. Significant variation in performance was observed across subsets, with the top-performing submission achieving a \qty{67.1}{\percent} gain for the HSN subset over random sampling and a gain of \qty{8}{\percent} for the ATBFL subset. Top-ranking submissions combined multiple acquisition signals, and diversity-based selection outperformed pure uncertainty sampling. Furthermore, there is evidence that transitioning from diversity-based to uncertainty-based selection and explicitly reducing redundancy within acquisition batches improve model training. There is also initial evidence that larger acquisition batch sizes may be increasingly beneficial later in the labelling process.

\end{abstract}

\section{Introduction}
\label{sec:intro}
A fundamental challenge across bioacoustics, in terrestrial and marine domains alike, is the cost of annotation. Passive acoustic monitoring (PAM) systems generate vast amounts of data \cite{cretois2026tabmon}, but only a small proportion can be feasibly annotated by human experts. Since model performance depends heavily on the quality and quantity of labelled data, this raises the following question: \textit{Given vast amounts of raw acoustic data and limited annotation resources, which data should be prioritised for labelling?}\\
 
Active learning (AL) is an iterative process of model-guided sample selection and label acquisition that aims to maximise model performance with fewer labelled samples than random (passive) sampling~\cite{settles2012active}. Practical constraints on annotator availability and computation mean that AL for bioacoustics is generally applied in batch mode over a pool of unlabelled recordings: an acquisition function defines a notion of utility for unlabelled instances, which informs batch selection before the model is retrained. Acquisition function design must therefore balance the informativeness of individual instances against redundancy within the selected batch. Increasingly, this selection operates not on audio but on embeddings from a frozen, pretrained acoustic (foundation) model such as BirdNET~\cite{kahl2021birdnet}, Bird-MAE~\cite{rauch2025canmasked}, and PerchV2~\cite{van2025perch}, with only a lightweight classification head trained during the AL loop. This \textit{active fine-tuning} regime is increasingly common in bioacoustics, rather than end-to-end fine-tuning~\cite{rauch2024deepactivelearningavian, kath2024leveraging, mcewen2024active, dumoulin2025search}. Bioacoustic applications additionally present severe class imbalance, sparse and rare events, and domain shift across spatial and temporal gradients~\cite{rauch2025birdset}.
 
The BioDCASE challenge~\cite{stowell2026biodcase} is an annual bioacoustics data challenge, within the long-running format of the DCASE (Detection and Classification of Acoustic Scenes and Events) challenge~\cite{stowell2015detection,mesaros2025decade}. The 2026 BioDCASE challenge comprised six concurrent tasks, including AL for Bioacoustics as Task~4. The challenge consists of a development phase, during which the participants are provided development datasets, instructions, and baseline methods from which to develop their method, followed by an evaluation and reporting phase where participant submissions are ranked.

Growing interest in AL methods within bioacoustics and biodiversity monitoring creates a need for standardised evaluation across datasets, domains, and metrics. Evaluation should also consider practical objectives such as computational and annotation cost. The challenge design including dataset and evaluation metric selection is reported in Section~\ref{sec:design}.
The challenge provides a consistent test suite for evaluation of AL methods through the development and use of BaseAL (discussed in Section~\ref{sec:baseal}). The format also provides an accessible entry-point to AL methods for non-practitioners.

\subsubsection*{Contributions.} We report the first edition of the 2026 AL for Bioacoustics task. We discuss the challenge design for terrestrial and marine bioacoustics, and the development of two curated AL datasets publicly available on Zenodo~\cite{kurinchi_vendhan_2026_19133112,rauch2026_dcasebirdset}. We report results for 14 sampling methods, including ten submissions from seven teams and four baselines. We analyse sampling methods on held-out test sets and identify common characteristics among the top-performing submissions, including combining multiple acquisition signals, shifting from diversity-based to uncertainty-based selection over the course of sampling, and explicitly reducing redundancy within acquisition batches. Additionally, we discuss broader considerations for future iterations of the challenge, such as the choice of pre-trained model and datasets, and balancing evaluation across broader objectives such as computational and annotation cost.

\section{Challenge Design}
\label{sec:design}

The BioDCASE challenge attract a heterogeneous participant base with students, early-career researchers, domain experts and newcomers from across domains (e.g. ecology, acoustics, computer science). The challenge design, discussed here, must balance the flexibility and evaluation rigor with accessibility.
 
\subsection{Datasets}
\label{sec:datasets}
The challenge includes both terrestrial and marine datasets to assess whether AL methods generalize across distinct bioacoustic monitoring settings. Despite differences in species, recording conditions, and acoustic structure, both domains share the challenge of selecting informative samples under limited annotation budgets. Table~\ref{tab:datasets} provides a corresponding overview of our challenge's datasets.

\subsubsection*{BirdSet.} The three terrestrial datasets are derived from the labelled PAM test subsets of BirdSet~\cite{rauch2026_dcasebirdset,rauch2025birdset}: \textit{High Sierra Nevada} (HSN), \textit{Powdermill Nature Reserve} (POW), and \textit{Hawai'i Island} (UHH). HSN~\cite{mary_clapp_2023_7525805} originates from recordings collected in 2015 around ten high-elevation lakes in Sequoia and Kings Canyon National Parks, California, POW~\cite{chronister_2022_4656848} from dawn chorus recordings collected in 2018 at Powdermill Nature Reserve in Pennsylvania, and UHH~\cite{amanda_navine_2022_7078499} from recordings acquired at four sites on Hawai'i Island between 2016 and 2022. In the BirdSet benchmark, models are trained on weakly labelled focal recordings and evaluated on strongly annotated soundscapes, introducing a domain shift between training and evaluation data. For this challenge, we instead derive new train, validation, and test splits exclusively from the soundscape recordings. The resulting setting is therefore not directly comparable to the BirdSet benchmark, but more closely reflects an AL scenario in which unlabelled and labelled samples originate from the same PAM domain.

BirdSet standardized the source recordings to \si{32}{kHz}, aligned species labels using the eBird taxonomy, and divided the soundscapes into \si{5}{s} segments. Each segment was propagated through PerchV2~\cite{van2025perch}, resulting in a 1{,}536-dimensional embedding. Species represented by fewer than three positive segments were removed because they could not be included in all three data splits, reducing the target space from 21 to 19 classes for HSN and from 48 to 41 for POW, while UHH retained all 25 classes. For each of the three datasets, we generated multiple multilabel-aware candidate splits and selected a splitting in which every retained class occurred in the train, validation, and test splits while keeping label distributions comparable. Splits were constructed at the level of individual \si{5}{s} segments rather than complete source recordings. Hence, segments from the same source recording can occur in different splits.

\subsubsection*{Acoustic Trends Blue Fin Library (ATBFL).}

ATBFL, the representative marine dataset for the challenge, is an annotated library of PAM recordings collected around Antarctica between 2005 and 2017, designed to support the development and evaluation of automated detectors for Antarctic blue whales (\textit{Balaenoptera musculus intermedia}) and fin whale (\textit{Balaenoptera physalus}) vocalizations \cite{miller2020annotated}. To capture diverse geographic, temporal, and instrumentation settings, recordings span 11 site--year deployments across the Atlantic, Pacific, and Indian sectors of the Southern Ocean and the Western Antarctic Peninsula \cite{miller2020annotated}.

For this challenge, we converted the strongly-annotated ATBFL recordings released for BioDCASE 2025 Task 2 into a segment-level multilabel classification dataset \cite{jean_labadye_2025_15092732}. The original annotations specify the call type and start and end time of each vocalization, and the absolute timestamps were first converted to offsets relative to the start of each recording. Recordings were then divided into consecutive, non-overlapping 5-second windows, and passed as input to PerchV2 \cite{van2025perch} to generate input representations for the challenge. Strong event annotations were converted to window-level multilabel targets by assigning each segment all call types whose annotated intervals overlapped with the 5-second window, and segments with no overlapping call were treated as negative examples.

The resulting dataset contains 19,633 audio samples, corresponding to approximately 21.7 hours of audio \cite{kurinchi_vendhan_2026_19133112}. Samples preserve the site--year metadata and are annotated in a multilabel setting (multiple call types to co-occur within a sample) for seven blue- and fin- whale call types: Antarctic blue whale A (\texttt{bma}), AB/B (\texttt{bmb}), and Z (\texttt{bmz}) stereotyped calls, blue whale frequency-modulated D-calls (\texttt{bmd}), and fin whale downsweeps (\texttt{bpd}), 20-Hz pulses (\texttt{bp20}), and 20-Hz pulses with additional higher-frequency energy (\texttt{bp20plus}) \cite{miller2020annotated}. Of the 19,633 samples, 15,667 (79.8\%) contain at least one target call, with substantial differences in prevalence across call types \cite{kurinchi_vendhan_2026_19133112}. Table \ref{tab:datasets} summarizes the train, validation, and test splits for this dataset.

\begin{table*}[!h]
    \setlength{\tabcolsep}{0.25cm}
    \centering
    \caption{Statistics of our challenge's datasets after preprocessing.}
    \label{tab:datasets}
    \begin{tabular}{lcccl}
    \toprule
    Subset & Segments & Classes & Labels per Segment & Train/Validation/Test \\
    \midrule
    HSN    &  12{,}000              & 19 & 0{.}52   & \phantom{0}6{,}600/1{,}800/\phantom{0}3,600     \\
    POW    &  \phantom{0}4{,}560    & 41 & 2{.}83   & \phantom{0}2{,}280/\phantom{0.}684/\phantom{0}1,596       \\
    UHH    &  36{,}637              & 25 & 1{.}05  & 18{,}319/7{,}327/10{,}991   \\
    ATBFL & 19{,}633 & 7 & 2{.}26 & 12{,}506/3{,}127/4{,}000 \\
    \bottomrule
    \end{tabular}
    \label{tab:placeholder}
\end{table*}

\newpage

\subsection{Representations and Model}


Participants received pre-computed PerchV2 \cite{van2025perch} embeddings (generated using bacpipe \cite{kather2026bacpipe}) rather than the raw audio, matching the active fine-tuning regime increasingly common in bioacoustics. PerchV2 was selected for its common usage within the bioacoustics community, benchmarking on marine mammal and underwater acoustics \cite{burns2025perch}, and competitive performance relative to other pre-trained models \cite{schwinger2026foundation}. This approach kept the datasets compact and the pipeline runnable on consumer-grade hardware without HPC access. The implications of this design are discussed in Section~\ref{sec:design-constraint}.

The classification head was fixed to a two-layer MLP (1536-dimension projection layer and output layer based on class count) to prevent trivial improvements unrelated to AL such as increasing the number of model parameters.

\subsection{BaseAL: Active Learning Framework}
\label{sec:baseal}

\begin{figure}[H]
    \centering
    \includegraphics[width=0.80\linewidth]{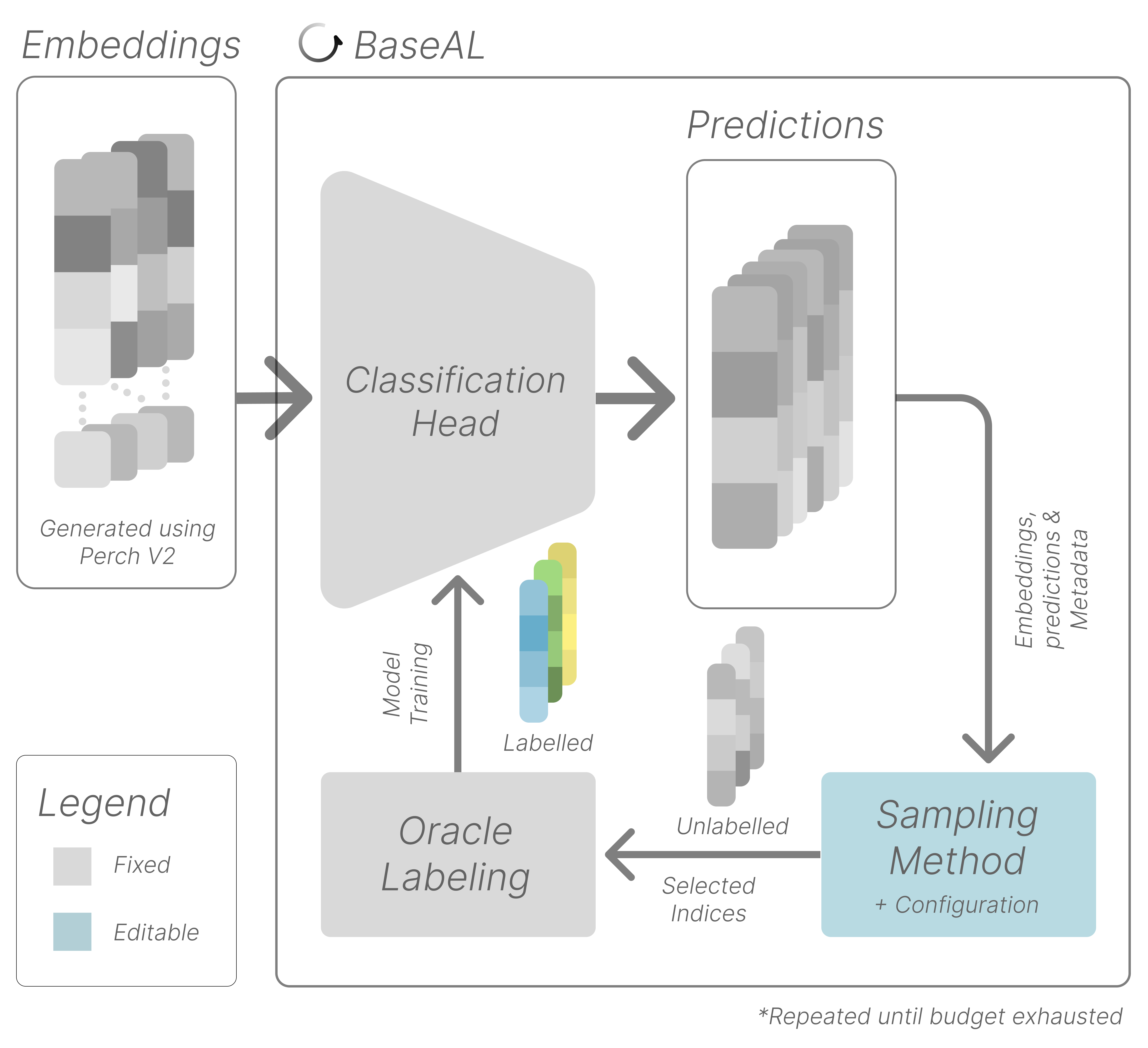}
    \caption{Active learning loop showing fixed and editable components.}
    \label{fig:AL}
\end{figure}

\href{https://github.com/BenMcEwen1/BaseAL}{BaseAL} is an AL framework designed for the BioDCASE Active Learning for Bioacoustics challenge \cite{mcewen2026_baseal}. BaseAL provides a general AL pipeline split into modular components, user-friendly notebooks, and an accompanying \href{https://baseal.up.railway.app/}{web-application} for visualising AL cycles in the embedding space and comparing sampling methods. Figure~\ref{fig:AL} shows the AL cycle and which components were fixed and which components participants could edit.

At each AL cycle the participant acquisition function received the current model predictions, the embeddings, dataset-specific metadata and the indices of the labelled and unlabelled sets. The function returns a utility score for each unlabelled sample. The highest-scoring samples were then revealed by oracle labelling, added to the labelled set and the classification head is retrained. This repeated until the annotation budget was exhausted.

Participant flexibility was bounded. Participants could edit the acquisition function and an optional warm-up function used for the first cycle before any labels are available. They could additionally set the AL batch size, and therefore the number of AL cycles. The core loop and experiment manager, the model, and the embeddings were fixed. The options to vary batch size and cycle counts affects computational cost, which is reported as a supplementary metric (Section~\ref{sec:evaluation}).

\subsection{Evaluation Protocol}
\label{sec:evaluation}
The first edition of the challenges only ranks participants on model training efficiency. The primary ranking metric was Area Under the Learning Curve (AULC) of  macro-averaged average precision (mAP). Learning curves were computed for participant submissions up to the maximum budget, submissions with a higher AULC mAP achieved a higher model performance within the budget. The final score was averaged over each of the four test data subsets, meaning that to achieve a high overall ranking, submissions must generalise (e.g. across locations or domains, marine vs. terrestrial). Test splits use the same stratified, label-aware procedure as train/validation (Section~\ref{sec:datasets}), so difficulty and label prevalence are comparable across splits.

A major consideration when applying AL is computational and annotation cost. For transparent ranking, these considerations were not included in the ranking score however they were reported as supplementary metrics. \textit{Computational cost} was measured as the sampling time (s), the number of seconds on average require to assign utility scores and select samples. Additionally the computational cost $cost\_method = model\_parameters * epochs * AL\_cycles$ relative to the baseline configurations $relative\_cost = cost\_method / baseline\_cost$. Participants could opt to increase the number of AL cycles (reduce the AL batch size) or increase the number of model training epochs. For large unlabelled datasets,repeated AL cycles can be extremely costly, leading to larger batch sizes. When using oracle labelling, \textit{annotation cost/time} cannot be measured directly. Instead a proxy of the total number of labels per samples was used. The intuition was that audio segments with more vocalising species would be more time-consuming to annotate. All metrics were averaged over subsets and over five independent repeats.
 
\subsection{Baselines and Configuration}
\label{sec:baselines}

Four baseline methods were provided to participants: random, margin (multilabel), CoreSet \cite{sener2017active} and TypiClust \cite{hacohen2022active}. Random sampling (passive sampling) simulates model training if active learning had not been applied. Margin is a common uncertainty sampling method. In the multilabel setting, for each sample, the distance of each class probability from 0.5 (highest uncertainty) is averaged across classes and the samples with the smallest margin are selected. Margin represents a pure uncertainty-based sampling method, while CoreSet is a pure diversity criterion without explicit consideration of prediction uncertainty. CoreSet applies a k-center (farthest-first) selection in an embedding space, seeded by the labelled set, to minimize the maximum distance of unlabelled samples to the selected samples. TypiClust is also a diversity-based method which selects typical samples (i.e. highest local density) across clusters. 

A fixed batch size of 50 samples, with 10 model training epochs per AL cycle and a learning rate of 1e-3 was used for all baselines, up to the maximum budget of 500 samples for each of the four data subsets. The maximum budget was selected by manually testing where model performance was starting to plateau for each subset. For simplicity this 500 samples was selected for each of the subsets. This budget simulates a realistic labelling budget. All baseline results and participant results were averaged over five independent runs and the baseline results are reported in Table~\ref{tab:baselines} along with supplementary metrics. CoreSet achieved the highest baseline performance with an AULC (mAP) of 0.460, compared to passive sampling which achieved a score of 0.390.

\begin{table}[H]
\centering
\caption{Baseline sampling method performance on the development set,
aggregated across all subsets. All baselines use the reference
configuration (batch size 50, 10 epochs per cycle, learning rate
$10^{-3}$, budget 500 samples). Best AULC in bold.}
\label{tab:baselines}
\begin{tabular}{lcccc}
\toprule
Method & AULC (mAP) & Comp. cost & Sampling time (s) & Annotation cost \\
\midrule
Random      & 0.390          & 1.0 & 0.00131          & 966.85           \\
Margin      & 0.399          & 1.0 & 0.00307          & 1267.75          \\
CoreSet     & \textbf{0.460} & 1.0 & 5.04364          & 1019.05          \\
TypiClust   & 0.423          & 1.0 & 5.90751          & 958.10           \\
\bottomrule
\end{tabular}
\end{table}
 
\subsection{Submissions}
\label{sec:submissions}

Submissions followed the general BioDCASE \href{https://biodcase.github.io/challenge2026/submission}{submission} format \cite{stowell2026biodcase}, with some modifications specific to the challenges setting. Rather than submitting model predictions, participants submitted the file containing their acquisition function and warm-up function in some cases. This enabled evaluation of the acquisition function on a held-out test set and that the supplementary evaluation metric, sampling time, could be compared on the same hardware. For reproducibility, participants also provide a \textit{.yaml} file containing their submissions output on the development set. The development results were reported to a running leaderboard during the development phase of the challenge. Participants were also required to include a short technical report (max. 5 pages), summarising their method and design consideration.

Optionally, participants could note any additional dependencies not already included in BaseAL. An optional notebook was also recommended for participants using non-default configurations or AL batchsize schedulers. As per the evaluation instructions, both participant-generated results and organiser-generated results on the test set were repeated over five independent runs, and the average was reported. Participant submissions were uploaded to the BioDCASE organiser managed Microsoft Conference Management Toolkit (CMT). 

\section{Results}
\label{sec:results}

Fourteen sampling methods were ranked, comprised of ten participant submission and four baseline methods. Submissions were received from 12 academic institutes and companies spanning nine countries. Here we report the overall ranking (Section~\ref{sec:ranking}), performance across the four subsets (Section~\ref{sec:per_subset}) and the awarded submissions (Section~\ref{sec:awards}). The official results are available on the BioDCASE \href{https://biodcase.github.io/challenge2026/task4-results}{website}.

\subsection{Overall Ranking}
\label{sec:ranking}

Note that the following results are reported on the held out test dataset. Reported results will differ from those reported in the participant \href{https://biodcase.github.io/challenge2026/task4-results#technical-reports}{technical reports}.

\begin{table}[H]
\centering
\footnotesize
\caption{Task~4 submission ranking. Score is the AULC of macro mAP
averaged across the four evaluation subsets. Efficiency metrics are
reported but not used for ranking. Baseline entries are italicised.}
\label{tab:ranking}
\resizebox{0.95\textwidth}{!}{
\begin{tabular}{cllccccc}
\toprule
Rank & Authors & System & Report & Score & Cost & Sampling (s) & Labels \\
\midrule
1  & Dubus et al.          & ADU-MMR                 & \cite{dubusAdaptive2026} & 0.507 & 2.0 & 1.006 & 1026.7 \\
2  & Magaldi et al.        & CARE-DPP                & \cite{magaldiCARE2026} & 0.505 & 1.1 & 0.215 & 1124.6 \\
3  & Wang et al.           & PB-MFS                  & \cite{wangPBMFS2026} & 0.499 & 1.6 & 0.485 & 1109.9 \\
4  & Yang et al.           & Safe Rarity             & \cite{yangSafeRarity2026} & 0.492 & 1.0 & 0.385 & 1116.0 \\
5  & Nihal et al.      & Capped Rarity K-Center  & \cite{nihalRarity2026} & 0.481 & 0.9 & 0.128 & 1134.5 \\
6  & Garcia-Yi, J.      & AFL                     & \cite{garciaYiAFL2026} & 0.477 & 1.6 & 1.333 & 1031.5 \\
7  & Parcerisas et al.     & Coreset KMeans          & \cite{parcerisas2026} & 0.469 & 1.0 & 4.095 & 1012.3 \\
8  & \textit{Baseline}  & \textit{CoreSet}        & - & 0.464 & 1.0 & 5.838 & 1023.6\\
9  & Parcerisas et al.     & Coreset Eigenvalues     & \cite{parcerisas2026} & 0.435 & 0.8 & 7.800 & 1010.6 \\
10 & \textit{Baseline}  & \textit{TypiClust}      & - & 0.421 & 1.0 & 5.076 & 959.0 \\
11 & \textit{Baseline}  & \textit{Margin}         & - & 0.408 & 1.0 & 0.002 & 1263.7 \\
12 & \textit{Baseline}  & \textit{Random}         & - & 0.401 & 1.0 & 0.001 & 966.1 \\
13 & Parcerisas et al.     & All Quantiles KMeans    & \cite{parcerisas2026} & 0.397 & 1.0 & 0.330 & 992.1 \\
14 & Parcerisas et al.     & Balance Class (Eigen.)  & \cite{parcerisas2026} & 0.393 & 0.8 & 0.110 & 1039.0 \\
\bottomrule
\end{tabular}}
\end{table}

\subsection{Per-Subset Performance}
\label{sec:per_subset}

System performance was evaluated across data subsets (Table~\ref{tab:subsets}). The relative gain of the best-performing submission over random sampling
varied substantially across subsets. The largest gain was 67.1\% on HSN
(0.625 versus 0.374), compared with 8.0\% on ATBFL
(0.502 versus 0.465), 13.2\% on POW (0.490 versus 0.433), and 28.5\%
on UHH (0.428 versus 0.333). Submission rankings on ATBFL were strongly
consistent with those obtained by averaging AULC across the three BirdSet
subsets (HSN, POW, and UHH; Spearman's $\rho=0.855$ across ten
submissions), although the rankings were not identical.

\begin{table}[H]
\centering
\footnotesize
\caption{Per-subset AULC (mAP). ATBFL is the marine dataset; HSN, POW
and UHH are BirdSet subsets. Best value per subset in bold. Baseline
entries are italicised.}
\label{tab:subsets}
\begin{tabular}{clcccccc}
\toprule
\multirow{2}{*}{Rank} & \multirow{2}{*}{System} & \multirow{2}{*}{Score}
 & Marine & \multicolumn{3}{c}{Terrestrial (BirdSet)} \\
\cmidrule(lr){4-4}\cmidrule(lr){5-7}
 & & & ATBFL & HSN & POW & UHH \\
\midrule
1  & ADU-MMR                 & 0.507 & \textbf{0.502} & \textbf{0.625} & 0.480          & 0.421          \\
2  & CARE-DPP                & 0.505 & 0.500          & 0.601          & \textbf{0.490} & \textbf{0.428} \\
3  & PB-MFS                  & 0.499 & 0.491          & 0.601          & 0.481          & 0.422          \\
4  & Safe Rarity             & 0.492 & 0.492          & 0.589          & 0.475          & 0.412          \\
5  & Capped Rarity K-Center  & 0.481 & 0.489          & 0.568          & 0.474          & 0.392          \\
6  & AFL                     & 0.477 & 0.483          & 0.552          & 0.462          & 0.411          \\
7  & Coreset KMeans          & 0.469 & 0.496          & 0.555          & 0.440          & 0.386          \\
8  & \textit{CoreSet}        & 0.464 & 0.477          & 0.531          & 0.453          & 0.394          \\
9  & Coreset Eigenvalues     & 0.435 & 0.457          & 0.484          & 0.435          & 0.364          \\
10 & \textit{TypiClust}      & 0.421 & 0.477          & 0.430          & 0.421          & 0.358          \\
11 & \textit{Margin}         & 0.408 & 0.457          & 0.438          & 0.419          & 0.319          \\
12 & \textit{Random}         & 0.401 & 0.465          & 0.374          & 0.433          & 0.333          \\
13 & All Quantiles KMeans    & 0.397 & 0.461          & 0.402          & 0.393          & 0.332          \\
14 & Balance Class (Eigen.)  & 0.393 & 0.447          & 0.389          & 0.408          & 0.328          \\
\midrule
\multicolumn{3}{l}{\textit{Range across all entries}}
   & 0.055 & 0.251 & 0.097 & 0.109 \\
\bottomrule
\end{tabular}
\end{table}

\subsection{Awards and Submissions}
\label{sec:awards}

All ten systems were hybrid (diversity- and uncertainty-based) acquisition functions. Every team placed at least one entry above the strongest baseline method (CoreSet, 0.464). 

\subsubsection*{Top Ranking.}
ADU-MMR (Dubus, Magaldi and Gros-Martial \cite{dubusAdaptive2026}) ranked first with a AULC mAP score of $0.507 \pm 0.001$ and was the best entry on two of the four subsets (ATBFL, 0.502; HSN, 0.625). Each sample is scored as a weighted sum of per-class binary entropies with the normalised squared distance to the nearest labelled neighbour. The weighting parameter $a_t$ is a function of the classifiers confidence over the unlabelled pool. Uncertainty contributes nothing until the mean entropy falls below a threshold $\tau = 0.05$ and its weight is capped at 0.5 so the diversity term never carries less weight than the uncertainty term. Batches are then diversified by greedy Maximum Marginal Relevance over the l2-normalised embeddings with $\lambda = 0.3$. The team provides an ablation of adaptive weighting compared to fixed settings demonstrating clear improvement using adaptive weighting. ADU-MMR achieved an average annotation cost of 1026.7, comparable to the CoreSet baseline. The relative cost was 2x higher than the baseline configuration, with a reduced acquisition batch of 25 samples (compared to 50), doubling the number AL cycles. This was the highest cost of all entries. The average sampling time was 1.006 s.

CARE-DPP (Magaldi and Dubus \cite{magaldiCARE2026}) ranked second at $0.505 \pm 0.007$ AULC mAP and was the best entry on the subsets POW (0.490) and UHH (0.428). The averaged ranking score across subsets was only 0.002 below ADU-MMR with an average SD of 0.007 meaning that the performance gap between the first and second ranking submissions are not statistically significant. CARE-DPP has a relative cost of only 1.1x higher than the baseline configuration and a sampling time of 0.215 s. This method uses a similar weighting between uncertainty and diversity but instead anneals based on budget rather than the model state. Novelty weight is dropped from 0.65 to 0.25 at 250 labels and uncertainty weight rising comparatively. CARE-DPP selects batches using a determinantal point process (DPP). The ablation notes a reduction in mean AULC of 0.038 (0.502 to 0.464) when removing DPP. This is a significant reduction relative to the other scores.

\subsubsection*{Jury Award.}
PB-MFS (Wang et al. \cite{wangPBMFS2026}) ranked third at 0.499 and was selected for the novelty of their candidate filtering method and the completeness of the technical report. This method decouples filtering from selection. Four filters of ordered criteria: coverage, mean-margin uncertainty, class-frequency and disagreement with the K nearest labelled neighbours and label cardinality, which successively prune the pool at a retention ratio of 0.3 per stage. The votes based on each of the filters and each sample is ranked. The optimal annotation batch then balances uncertainty and diversity through a Pareto optimisation objective. The relative cost of this method was 1.6x the baseline configuration and the average sampling time was 0.485 s.

\subsubsection*{Notable Submissions}
Safe Rarity (Yang \cite{yangSafeRarity2026}, rank 4, 0.492) mixes rank-normalised top-5 binary entropy, k-center diversity and a predicted-prevalence rarity term, with the rarity weight passed through a gate that collapses to zero when predictions are degenerate or the budget is small, reducing the sampler to entropy-filtered k-center. It achieved the best rank per unit compute of any entry, placing fourth at a relative cost of 1.0 and a sampling time of 0.385 s. Its ablation attributes the largest effect to k-center diversity. Capped Rarity K-Center (Nihal et al. \cite{nihalRarity2026}, rank 5, 0.481) is the only submission to use dataset metadata, adding a location-stratification term alongside a capped class-deficit rarity term and a budget-ramped margin computed from a kNN label surrogate, with all three faded in or out by a single density gate keyed to the average number of labels per sample. AFL (Garcia-Yi \cite{garciaYiAFL2026}, rank 6, 0.477) is the most distinctive, using a discrete particle-swarm search over candidate subsets for its 50-sample warm start and a front-loaded budget schedule, then coverage-only selection until 35\% of the budget is spent before switching to a hybrid score. Its ablations rank the swarm warm start and diversity-aware batch reranking as the two largest contributors (-0.014 each), while the front-loaded scheduler and the density term were each worth under 0.001. The Parcerisas et al. \cite{parcerisas2026} submitted four ranked entries and is the only team to have compared warm-up procedures systematically. Their entries ranked as Coreset KMeans placed 7th (0.469), above the CoreSet baseline, while Coreset Eigenvalues placed 9th (0.435) and their two confidence-quantile strategies placed 13th and 14th, below random sampling.

\section{Discussion}
\label{sec:discussion}
 
\subsection{Progress Towards Active Learning for Bioacoustics}
\label{sec:progress}
The challenge provides further evidence that AL can improve label efficiency for bioacoustic classification. An earlier pilot study on HSN found that diversity-based sampling was particularly effective early in the AL process, while uncertainty became more useful once additional labels had been acquired~\cite{rauch2024deepactivelearningavian}. A similar pattern emerged across the independent submissions in this challenge. CoreSet was the strongest baseline, outperforming both random and margin-based uncertainty sampling, while the highest-ranked submissions generally retained an explicit coverage or diversity component and combined it with uncertainty \cite{dubusAdaptive2026, magaldiCARE2026, wangPBMFS2026}, rarity \cite{yangSafeRarity2026, nihalRarity2026}, or class-balance information \cite{magaldiCARE2026, wangPBMFS2026}. Several methods further adapted these signals over the annotation budget \cite{magaldiCARE2026, garciaYiAFL2026} or according to model confidence \cite{dubusAdaptive2026, yangSafeRarity2026, nihalRarity2026}. These results suggest that coverage of the representation space provides a robust foundation for acquisition when labels are scarce in bioacoustics, while model-dependent information becomes more useful as the classifier improves.

The benefit of AL varied substantially across subsets. Relative to random sampling, the best submission improved AULC by 67.1\% on HSN and 28.5\% on UHH, compared with 13.2\% on POW and 8.0\% on ATBFL (\autoref{tab:subsets}). These differences indicate that the potential gains from AL depend strongly on characteristics of the underlying dataset. HSN, the sparsest BirdSet subset, showed the largest separation between methods, although the small number of subsets prevents attributing this effect to label density alone. Overall, the results support combining coverage with adaptively introduced model-based criteria rather than relying on a single acquisition method across the AL process.


Two teams varied acquisition batch size in opposite directions: CARE-DPP \cite{magaldiCARE2026} increased it as labels accumulated (25 to 75), for a modest 0.014 gain, while AFL's \cite{garciaYiAFL2026} ablation with larger early-cycle batches showed no improvement. The leading submission instead used a smaller fixed batch (25 vs. 50), doubling AL cycles and computational cost. Optimising batch size across the budget, or extending AL to larger batches \cite{citovsky2021batch}, remains a valuable direction given the cost of repeated inference for large PAM deployments \cite{cretois2026tabmon, bernard2026data}.

Parcerisas et al. \cite{parcerisas2026} systematically compared warm-up methods across ten acquisition strategies finding little effect (0.003 AULC difference) compared to that of the acquisition function (with a difference 0.121). Two other participants found a minor improvement in performance using warm-up with separate ablations showing a 0.026 \cite{wangPBMFS2026} and 0.014 \cite{garciaYiAFL2026} increase in performance on the development set. Further investigation is required to determine the utility of warm-up in the case of high-quality pre-trained model embeddings applied within similar domains (bioacoustics-to-bioacoustics). 

\subsection{Single Metric or Aggregate}
\label{sec:aggregate}
The leaderboard is based on a single metric in the form of AULC scores for macro mAP aggregated across datasets. While this provides a straightforward scalar ranking of the AL methods, there are several limitations: AULC is not directly comparable across datasets with different difficulty levels and label prevalences, a single metric captures only a single aspect of the systems performance. Finally the current ranking does not consider practical considerations such as segment-level labelling cost and the computational cost of different AL acquisition functions and configurations.
Several of these limitations are implicitly considered by including proxies for them as separate statistics in Table~\ref{tab:ranking}. This leads to the question whether we can improve upon this by aggregating multiple metrics. One option would be to consider rank aggregation across datasets and evaluation criteria, as proposed for multi-task and multi-metric benchmarking~\cite{colombo2022what}. These ranks could be complemented with normalized performance differences, where we, for example, employ random sampling as lower and an oracle AL method~\cite{huseljic2026boss} as upper references. Further additions may consider the actual segment-dependent labelling costs when computing AULC and incorporate computational efficiency either as a separate evaluation criterion or by expressing labelling and compute costs on the same scale. Estimating the labelling costs may leverage the number of labels per segment as a first crude proxy but requires further validation. A potentially supportive observation for such a proxy is that Margin, a purely exploitative AL method, selected on average segments with more labels than the remaining AL methods (cf.\ Table~\ref{tab:ranking}). 
 
\subsection{Configuration, Generalisability and Deployability}
\label{sec:hyperparams}

Label scarcity motivates AL but also makes acquisition configurations difficult to validate. Comparing settings often requires labelled validation data and repeated AL runs. L\"uth \etal describe the AL validation paradox: configuration selection can consume the labels that AL aims to save \cite{lueth2023navigating}. Past acquisition decisions cannot be revised once their labels have been collected \cite{lowell2019practical}, and gains over random sampling may diminish under carefully controlled training and evaluation conditions \cite{munjal2022towards}. Here, ``deployability'' means applying a complete acquisition policy to a new data source without label-intensive retuning.

Submitted configurations used fixed weights, schedules, gates, candidate-pool rules, or warm starts. We did not count raw hyperparameters, as this conflates choices with different roles.
The submission format also did not standardise how values were chosen or whether development data informed them. The challenge fixed the classification head but allowed the acquisition configuration, query schedule, and warm start to vary (Section~\ref{sec:baseal}). The warm start directly affects the learning curve by determining the first labelled set included when the curve is integrated. Because the held-out evaluation data came from the same datasets and subsets represented during development, the ranking measures performance only on unseen samples within the predefined challenge domains, not transfer to a new site or data source.

Future challenge editions could test generalisability more directly by freezing each complete submitted configuration and applying it unchanged to source-disjoint evaluation data unavailable during development. They could also require standardised reports of configuration defaults, rationale, data dependencies, and tuning procedures. Rules relative to the annotation budget or properties of the unlabelled pool may accommodate changes in dataset scale, but their transfer would still require evaluation. Sensitivity analyses could identify which choices affect performance. Hyperparameter counts could remain descriptive metadata, but not stand-alone measures of generalisability or deployability.

\subsection{Foundation-Model Embeddings as a Design Constraint}
\label{sec:design-constraint}
For this iteration of the challenge, we provided all participants with fixed PerchV2 embeddings. Fixing the representations allowed for a controlled comparison of acquisition functions by preventing gains from being driven by differences in model training and architecture or output representation quality. The choice of foundation model itself affects AL performance, since many acquisition functions depend directly or indirectly on the geometry of the learned representations: changing the embedding model can change which samples are most uncertain redundant, representative, or distant from the labelled set. Dumoulin et al. 2025 compare multiple acoustic embedding models within an ``agile modeling'' workflow and find substantial differences in downstream classification performance across representations. Domain-relevant bioacoustic embeddings generally outperforming more generic audio representations 
\cite{dumoulin2025search}, with disproportionate effects for rare or easily-confused classes \cite{kurinchi2026finding}. Future iterations of this challenge could explicitly explore this dimension further by including comparisons of the initial representations, or ensembles of them.

\subsection{Dataset Selection and Representation Limits}
\label{sec:dataset-limits}
The ATBFL recordings are sampled at 250 Hz and contain predominantly low-frequency blue- and fin-whale vocalizations: several target calls have most of their energy below 60 Hz, including blue whale Z-calls and fin whale 20 Hz pulses \cite{miller2020annotated}. However, the PerchV2 frontend operates on a log-mel representation spanning approximately 60 Hz -- 16 kHz \cite{van2025perch}. 
Although some ATBFL calls contain higher-frequency components useful for discrimination, much of the signal energy falls below the PerchV2 range. However, this representation mismatch didn't prevent strong performance on ATBFL (the results summarized in Sec. \ref{sec:per_subset}). This suggests that PerchV2 embeddings are still able to discriminate between the call types. This may be because some calls contain harmonics, broadband structure, or high-frequency components above the 60 Hz frontend cutoff. The embeddings may also capture other acoustic characteristics correlated with call identity.

While future iterations of this challenge should continue to include both terrestrial \textit{and} marine datasets to test cross-domain performance, it may be more valuable to replace ATBFL with an underwater acoustics dataset that better matches existing embedding models. If PerchV2 remains a common representation for bioacoustics, datasets with stronger kilohertz-range signals, such as the DCLDE killer whale dataset \cite{dfo2025dclde}, would allow for cleaner comparisons of acquisition strategies. 
 
\subsection{Joint Allocation of the Annotation Budget}

An important consideration for deploying ML models for bioacoustic monitoring is validating model performance to support downstream ecological inference. While the sampling objective for validation differs from that of AL for model training, both draw from the same finite labelling budget and cannot, in most scenarios, be used interchangeably without introducing bias \cite{kurinchi2026finding, farquhar2021statistical, lowell2019practical}. This highlights a key limitation of the current data challenge format and evaluation of AL systems more generally: oracle sampling assumes a pre-existing validation set that is excluded from budget considerations. BaseAL V1.2 \cite{mcewen2026_baseal} now supports active testing, enabling iterative validation strategies within a budget. Future editions could leverage this to require participants to jointly optimise model performance and evaluation within a single annotation budget, better reflecting deployment scenarios.




\section{Conclusions}
\label{sec:conclusions}

The 2026 BioDCASE Active Learning for Bioacoustics challenge demonstrates that AL can substantially improve label efficiency across bioacoustics domains, with combining multiple acquisition signals, shifting from diversity- to uncertainty-based selection as labels accumulate, and explicitly reducing batch redundancy each improving over single-criterion methods. Future work should extend evaluation to better reflect deployment: systematic comparison of embedding models, computational and annotation cost, configuration generalisability, joint allocation of training and validation budgets, and coverage of more diverse taxa (anurans, bats, fish). The challenge format and BaseAL provide a foundation for continued progress in data-efficient bioacoustic monitoring.

\section*{Acknowledgements}
Thank you to the participants of the 2026 BioDCASE Active Learning for Bioacoustics challenge and to the BioDCASE organisers.

%
%
\bibliographystyle{splncs04}
\bibliography{main}
\end{document}